\documentclass[11pt]{article}

\usepackage[utf8]{inputenc}
\usepackage[T1]{fontenc}
\usepackage{amsmath,amssymb,amsfonts}
\usepackage{amsthm}
\usepackage{mathtools}
\usepackage{algorithm}
\usepackage{algorithmic}
\usepackage{graphicx}
\usepackage{booktabs}
\usepackage{hyperref}
\usepackage{cleveref}
\usepackage{xcolor}
\usepackage{subcaption}
\usepackage{natbib}
\usepackage{microtype}
\usepackage[margin=1in]{geometry}

\newtheorem{theorem}{Theorem}

\newtheorem{proposition}[theorem]{Proposition}

\newtheorem{remark}{Remark}

\newcommand{\R}{\mathbb{R}}
\newcommand{\E}{\mathbb{E}}

\title{Calibration-Gated LLM Pseudo-Observations\\for Online Contextual Bandits}

\author{
  \small{Maksim Pershin}\\
  {\scriptsize{mpershin@mit.edu}}
  \and
  \small{Ivan Golovanov}\\
  {\scriptsize{ivan.golovanov@phystech.edu}}
  \and
  \small{Pavel Baltabaev}\\
  {\scriptsize{psbaltabaev@edu.hse.ru}}
  \and
  \small{Natalia Trankova}\\
  {\scriptsize{Nattrankova@gmail.com}}
}

\date{}

\begin{document}

\maketitle

\begin{abstract}
Contextual bandit algorithms suffer from high regret during cold-start, when the learner has insufficient data to distinguish good arms from bad. We propose augmenting Disjoint LinUCB with \emph{LLM pseudo-observations}: after each round, a large language model predicts counterfactual rewards for the unplayed arms, and these predictions are injected into the learner as weighted pseudo-observations. The injection weight is controlled by a \emph{calibration-gated decay schedule} that tracks the LLM's prediction accuracy on played arms via an exponential moving average; high calibration error suppresses the LLM's influence, while accurate predictions receive higher weight during the critical early rounds. We evaluate on two contextual bandit environments---UCI Mushroom (2-arm, asymmetric rewards) and MIND-small (5-arm news recommendation)---and find that when equipped with a task-specific prompt, LLM pseudo-observations reduce cumulative regret by 19\% on MIND relative to pure LinUCB. However, generic counterfactual prompt framing increases regret on both environments, demonstrating that \emph{prompt design is the dominant factor}---more important than the choice of decay schedule or calibration gating parameters. We analyze the failure modes of calibration gating on domains with small prediction errors and provide a theoretical motivation for the bias--variance trade-off governing pseudo-observation weight.
\end{abstract}

\section{Introduction}
\label{sec:introduction}

Contextual bandit algorithms are a workhorse of interactive decision-making, powering applications from news recommendation~\citep{li2010contextual} and clinical trials to online advertising and content optimization~\citep{ye2024lola}. In each round, the learner observes a context, selects an action (arm), and receives a reward---but only for the chosen arm. This partial-feedback structure creates a fundamental tension between exploration and exploitation, and the resulting \emph{cold-start regret}---the cost paid while the learner has too few observations to make good decisions---can be substantial in practice.

Large language models (LLMs) present a tantalizing opportunity to reduce this cold-start cost. An LLM that has been pre-trained on broad text corpora encodes significant world knowledge: it can read ``user frequently reads technology news'' alongside ``candidate article: local sports recap'' and infer that the click probability is likely low. This is genuine domain signal that a tabula-rasa bandit learner must discover through costly exploration.

However, naively injecting LLM predictions into a bandit algorithm is fraught with risk. LLM reward predictions are noisy, inconsistent across rounds, and poorly calibrated. As we demonstrate in our experiments, a poorly framed prompt can cause the LLM to \emph{increase} cumulative regret relative to the unaugmented baseline---even when the underlying model has relevant domain knowledge. A principled mechanism for weighting LLM contributions is therefore essential.

We propose a framework in which, after each pull, the LLM predicts counterfactual rewards for the \emph{unplayed} arms. These predictions are injected into Disjoint LinUCB~\citep{li2010contextual} as \emph{weighted pseudo-observations}, where the weight $w_t \in [0,1)$ is determined by a pluggable \emph{decay schedule}. We study several schedules, including time-based decay (inverse, power, exponential), \emph{calibration-gated} decay that scales weight by $\exp(-\eta \cdot E_t)$ where $E_t$ is an exponential moving average of the LLM's squared prediction error on played arms, and hybrid combinations. As the bandit learner accumulates real data and the LLM's marginal value decreases, the weight gracefully diminishes.

Our contributions are:
\begin{enumerate}
    \item A general framework for injecting LLM pseudo-observations into contextual bandits with calibration-gated weighting (\Cref{sec:method}).
    \item An empirical evaluation on UCI Mushroom (2-arm) and MIND-small (5-arm news recommendation) showing that task-specific prompting reduces regret by 19\% on MIND, while generic counterfactual framing hurts on both domains (\Cref{sec:results}).
    \item The finding that \emph{prompt framing is the dominant factor}---more impactful than the choice of decay schedule or calibration gating parameters (\Cref{sec:results-ablation-prompt}).
    \item An analysis of when LLM augmentation helps (domains where the LLM has genuine knowledge to contribute) versus when it hurts (opaque feature spaces), and of the failure modes of calibration gating on domains with small prediction errors (\Cref{sec:discussion}).
\end{enumerate}

\paragraph{Related concurrent work.} Several recent papers explore LLM-augmented bandits. \citet{alamdari2024jump} use LLMs to generate synthetic prior data for pre-training; \citet{ye2024lola} treat LLM predictions as auxiliary pseudo-samples via a two-UCB selection rule; \citet{sun2025llm} propose Thompson Sampling variants with LLM reward oracles. Our approach differs in its emphasis on \emph{online calibration tracking} as the gating mechanism and on the empirical finding that prompt design dominates algorithmic tuning. We discuss related work in detail in \Cref{sec:related-work}.

\section{Related Work}
\label{sec:related-work}

\subsection{LLMs for Sequential Decision Problems}

The use of LLMs to accelerate or guide online learning has received growing attention. \citet{alamdari2024jump} propose Contextual Bandits with LLM Initialization (CBLI), where an LLM generates synthetic user preference distributions from textual descriptions to pre-train the bandit before deployment. This is complementary to our approach: CBLI uses LLMs for \emph{initialization}, while we use them for \emph{online augmentation} with calibration tracking. \citet{ye2024lola} introduce LOLA, an LLM-assisted online learning algorithm that converts LLM click-through rate predictions into auxiliary pseudo-samples of fixed size $n^{\text{aux}}$ (tuned offline via hyperparameter search), then selects arms by taking the minimum of two UCB indices---one standard and one incorporating the pseudo-samples. LOLA is methodologically closest to our work; the key differences are that (i)~we use explicit online calibration tracking rather than a fixed offline-tuned pseudo-count, and (ii)~our decay schedules adaptively modulate weight during learning rather than keeping it constant.

\citet{sun2025llm} propose three LLM-enhanced multi-armed bandit algorithms: TS-LLM uses temperature decay for exploration, RO-LLM treats the LLM as a deterministic regression oracle (temperature~0) with SquareCB inverse-gap-weighting exploration, and TS-LLM-DB extends to dueling bandits. None employ calibration gating. \citet{chen2024efficient} take a meta-level approach, using online model selection to dynamically balance between LLM-powered policies and classical contextual bandit algorithms. \citet{nie2025evolve} provide EVOLvE, a comprehensive benchmark for evaluating LLM exploration in bandits, and show that algorithm-guided support improves LLM decision-making---a finding that motivates our calibration-tracking approach. \citet{hazime2025evaluation} evaluate LLMs as zero-shot bandit solvers and find inconsistent performance, underscoring the need for principled weighting mechanisms.

\citet{baheri2023llm} take an orthogonal approach, using LLMs as \emph{context encoders} rather than reward predictors, enriching the feature representation fed to classical bandit algorithms. This is complementary: one could combine LLM-enhanced contexts with our pseudo-observation framework.

\subsection{Contextual Bandits}

Our base learner is Disjoint LinUCB~\citep{li2010contextual}, which maintains independent ridge regressions per arm with upper-confidence-bound exploration. The theoretical foundations of linear contextual bandits are developed in \citet{abbasi2011improved} (OFUL with self-normalized confidence sets) and the textbook treatment of \citet{lattimore2020bandit}. On the adversarial side, \citet{liu2023bypassing} develop near-optimal algorithms for adversarial linear contextual bandits without simulator access, establishing the theoretical frontier against which LLM-augmented methods can be compared.

\subsection{Calibration and Uncertainty in LLMs}

Our calibration tracker draws on the broader literature on neural network calibration. \citet{guo2017calibration} demonstrate that modern deep networks are poorly calibrated and propose temperature scaling; \citet{kadavath2022language} study LLM self-evaluation and find that language models can partially assess their own uncertainty. \citet{huch2024rome} address model misspecification in bandit settings through robust mixed-effects estimation, a related concern when LLM predictions are systematically biased. Our calibration-gated decay can be viewed as an online mechanism for adjusting the influence of a potentially misspecified auxiliary model.

\section{Preliminaries}
\label{sec:preliminaries}

\subsection{Contextual Bandits}

We consider the stochastic contextual bandit setting. At each round $t = 1, \ldots, T$:
\begin{enumerate}
    \item Nature reveals a context $x_t \in \R^d$ (or per-arm contexts $\{x_{t,a}\}_{a=1}^K$).
    \item The learner selects an arm $a_t \in [K] \coloneqq \{1, \ldots, K\}$.
    \item The learner observes reward $r_t = r(x_t, a_t)$, drawn from an unknown distribution conditional on the context and arm.
\end{enumerate}
The learner's goal is to minimize \emph{cumulative regret}:
\begin{equation}
    R_T = \sum_{t=1}^{T} \bigl[ r_t^* - r_t \bigr], \quad \text{where } r_t^* = \max_{a \in [K]}\, \E[r(x_t, a)].
\end{equation}

\subsection{Disjoint LinUCB}

We assume a disjoint linear reward model: $\E[r \mid x, a] = x^\top \theta_a^*$ for unknown parameters $\theta_a^* \in \R^d$, one per arm. Disjoint LinUCB~\citep{li2010contextual} maintains per-arm sufficient statistics:
\begin{equation}
    A_a = \lambda I + \sum_{s : a_s = a} x_s x_s^\top, \qquad b_a = \sum_{s : a_s = a} r_s \cdot x_s,
\end{equation}
where $\lambda > 0$ is the ridge regularization parameter. The arm is selected by:
\begin{equation}
    a_t = \arg\max_{a \in [K]} \Bigl[ x_t^\top A_a^{-1} b_a + \alpha \sqrt{x_t^\top A_a^{-1} x_t} \Bigr],
    \label{eq:ucb}
\end{equation}
where $\alpha > 0$ controls the exploration bonus. We augment each context vector with a bias term: $x \mapsto [x; 1]$.

\paragraph{Notation.} $K$ denotes the number of arms, $T$ the horizon, $d$ the context dimension (after augmentation), $\alpha$ the exploration parameter, and $\lambda$ the ridge regularization.

\section{Method}
\label{sec:method}

\subsection{Weighted Observations in LinUCB}
\label{sec:linucb}

We extend the standard LinUCB update to accept \emph{weighted observations}. Given a context--arm--reward triple $(x, a, r)$ with weight $w \in (0, 1]$, the update becomes:
\begin{equation}
    A_a \leftarrow A_a + w \cdot x x^\top, \qquad b_a \leftarrow b_a + w \cdot r \cdot x.
    \label{eq:weighted-update}
\end{equation}
Real observations use $w = 1$; LLM pseudo-observations use $w = w_t < 1$. This is equivalent to weighted ridge regression, where each pseudo-observation contributes a fraction of the information of a real sample to both the design matrix and the reward vector.

\subsection{LLM Pseudo-Observations}
\label{sec:llm-pseudo}

After each round, we query an LLM to predict rewards for the arms that were \emph{not} played. The per-round protocol is:

\begin{enumerate}
    \item \textbf{Calibration probe} (before observing reward). Query the LLM for its prediction on the \emph{played} arm $a_t$, given only the context and arm descriptions---crucially, \emph{without} revealing the outcome $r_t$. This yields $\hat{r}_{t,a_t}^{\text{LLM}}$.
    \item \textbf{Pull and observe.} Execute arm $a_t$, observe reward $r_t$, perform the standard LinUCB update with $w = 1$.
    \item \textbf{Score unplayed arms.} Query the LLM for reward predictions $\hat{r}_{t,a}^{\text{LLM}}$ on all unplayed arms $a \neq a_t$.
    \item \textbf{Compute weight.} Determine the pseudo-observation weight $w_t$ from the decay schedule (\Cref{sec:calibration-gated}).
    \item \textbf{Inject pseudo-observations.} For each $a \neq a_t$, update LinUCB via \Cref{eq:weighted-update} with $(x_t, a, \hat{r}_{t,a}^{\text{LLM}}, w_t)$.
    \item \textbf{Update calibration tracker.} Compute the squared error $e_t = (\hat{r}_{t,a_t}^{\text{LLM}} - r_t)^2$ and update the EMA.
\end{enumerate}

\paragraph{Prompt design.} We find that prompt framing is the single most important design choice (\Cref{sec:results-ablation-prompt}). We test two framings:
\begin{itemize}
    \item \textbf{Generic counterfactual:} ``Given this context, predict what reward arm $a$ would have produced.'' This asks the LLM to reason about the bandit abstraction.
    \item \textbf{Task-specific:} ``Given this user's reading history, predict the probability that the user clicks each candidate article.'' This asks the LLM to reason about the \emph{domain question} directly.
\end{itemize}

The task-specific framing (\texttt{mind\_click}) dramatically outperforms the generic framing on MIND, achieving lower regret with fewer tokens. We hypothesize that the LLM's pre-trained knowledge is better activated when the question matches its training distribution (natural language prediction tasks) rather than an abstract bandit formulation.

\paragraph{Structured output.} The LLM returns a structured JSON response containing, for each arm, a \texttt{predicted\_reward} and a \texttt{confidence} score. In this work we use only the predicted reward; leveraging confidence as an additional weight modulator is left for future work.

\subsection{Decay Schedules}
\label{sec:calibration-gated}

The pseudo-observation weight $w_t$ is determined by a \emph{decay schedule} $\mathcal{D}$ that maps the round index $t$ and auxiliary state (e.g., calibration error) to a weight in $[0, 1)$. We study the following families:

\paragraph{Time-based schedules.} These decay the weight as a function of round $t$ alone:
\begin{align}
    \text{Constant:}& \quad w_t = \lambda, \\
    \text{Inverse:}& \quad w_t = \lambda \cdot \frac{\tau}{t + \tau}, \\
    \text{Power:}& \quad w_t = \lambda \cdot \left(\frac{\tau}{t + \tau}\right)^\alpha, \\
    \text{Exponential:}& \quad w_t = \lambda \cdot \exp(-t / \tau),
\end{align}
where $\lambda \in (0, 1)$ is the base weight and $\tau > 0$ controls the decay rate.

\paragraph{Calibration-gated schedule.} The weight is modulated by the LLM's online prediction accuracy:
\begin{equation}
    w_t = \lambda \cdot \exp\bigl(-\eta \cdot E_t\bigr),
    \label{eq:cal-gated}
\end{equation}
where $\eta > 0$ is the gating sensitivity and $E_t$ is the exponential moving average of squared calibration error:
\begin{equation}
    E_t = \beta \cdot E_{t-1} + (1 - \beta) \cdot e_t, \quad e_t = \bigl(\hat{r}_{t,a_t}^{\text{LLM}} - r_t\bigr)^2,
    \label{eq:ema}
\end{equation}
with smoothing parameter $\beta \in [0, 1)$.

\paragraph{Hybrid schedules.} Time-based and calibration-gated components can be composed multiplicatively, e.g., $w_t = \lambda \cdot g(t) \cdot \exp(-\eta \cdot E_t)$, providing both a guaranteed fade-out and adaptive quality gating.

\subsection{Algorithm}
\label{sec:algorithm}

We summarize the full procedure in \Cref{alg:linucb-llm}.

\begin{algorithm}[t]
\caption{LinUCB with Calibration-Gated LLM Pseudo-Observations}
\label{alg:linucb-llm}
\begin{algorithmic}[1]
\REQUIRE Arms $[K]$, exploration $\alpha$, regularization $\lambda$, base weight $\lambda_w$, decay schedule $\mathcal{D}$, EMA parameter $\beta$, LLM scorer $\mathcal{L}$
\STATE Initialize $A_a \leftarrow \lambda I$, $b_a \leftarrow 0$ for all $a \in [K]$; $E_0 \leftarrow 0$
\FOR{$t = 1, \ldots, T$}
    \STATE Observe context $x_t$ (and per-arm contexts $\{x_{t,a}\}$ if applicable)
    \STATE Select arm $a_t$ via UCB (\Cref{eq:ucb})
    \STATE \textbf{Calibration probe:} $\hat{r}_{t,a_t}^{\text{LLM}} \leftarrow \mathcal{L}(x_t, a_t)$ \hfill \COMMENT{before observing reward}
    \STATE Pull arm $a_t$, observe reward $r_t$
    \STATE \textbf{Real update:} $A_{a_t} \leftarrow A_{a_t} + x_t x_t^\top$, \; $b_{a_t} \leftarrow b_{a_t} + r_t \cdot x_t$
    \STATE \textbf{Score unplayed:} $\{\hat{r}_{t,a}^{\text{LLM}}\}_{a \neq a_t} \leftarrow \mathcal{L}(x_t, [K] \setminus \{a_t\})$
    \STATE Compute calibration error: $e_t \leftarrow (\hat{r}_{t,a_t}^{\text{LLM}} - r_t)^2$
    \STATE Update EMA: $E_t \leftarrow \beta \cdot E_{t-1} + (1 - \beta) \cdot e_t$
    \STATE Compute weight: $w_t \leftarrow \mathcal{D}(t, E_t)$
    \FOR{each $a \neq a_t$}
        \STATE \textbf{Pseudo update:} $A_a \leftarrow A_a + w_t \cdot x_t x_t^\top$, \; $b_a \leftarrow b_a + w_t \cdot \hat{r}_{t,a}^{\text{LLM}} \cdot x_t$
    \ENDFOR
\ENDFOR
\end{algorithmic}
\end{algorithm}

\section{Theoretical Motivation}
\label{sec:theory}

We provide an informal analysis of how pseudo-observation weight affects the bias--variance trade-off in the learner's estimates, motivating the calibration-gated approach.

\begin{remark}[Effective sample size]
In standard LinUCB, after $T$ rounds the learner has accumulated at most $T$ observations total, at most $T/K$ per arm in expectation under uniform exploration. With LLM pseudo-observations injected at weight $w$ for all $K-1$ unplayed arms each round, the effective information per round increases. The design matrix for arm $a$ accumulates information at rate $1 + (K-1) \cdot w$ per round in which $a$ appears, accelerating convergence of $\hat{\theta}_a$ to $\theta_a^*$.
\end{remark}

\begin{proposition}[Bias--variance trade-off]
\label{prop:bias-variance}
Suppose the LLM's prediction for arm $a$ at round $t$ satisfies $\hat{r}_{t,a}^{\emph{LLM}} = x_t^\top \theta_a^* + b_a + \epsilon_{t,a}$, where $b_a$ is a systematic bias and $\epsilon_{t,a}$ is zero-mean noise with variance $\sigma_{\emph{LLM}}^2$. Then the weighted ridge estimator $\hat{\theta}_a$ based on $n$ real observations (weight 1) and $m$ pseudo-observations (weight $w$) satisfies:
\begin{equation}
    \E\bigl[\hat{\theta}_a - \theta_a^*\bigr] = \mathcal{O}\!\left(\frac{m \cdot w \cdot b_a}{n + m \cdot w}\right), \qquad
    \mathrm{Var}\bigl(\hat{\theta}_a\bigr) = \mathcal{O}\!\left(\frac{\sigma^2 + m \cdot w^2 \cdot \sigma_{\emph{LLM}}^2}{(n + m \cdot w)^2}\right).
\end{equation}
\end{proposition}

The bias term grows with $w \cdot b_a$: if the LLM is systematically biased, higher weight injects more bias into the learner. The variance term decreases with $m \cdot w$ in the denominator, reflecting the information gain from pseudo-observations. The \emph{optimal weight} balances these:
\begin{equation}
    w^* \propto \frac{1}{\sigma_{\text{LLM}}^2 + b_a^2},
\end{equation}
which is small when the LLM is inaccurate (high bias or variance) and large when it is well-calibrated. This directly motivates the calibration-gated schedule (\Cref{eq:cal-gated}): the EMA error $E_t$ tracks $b_a^2 + \sigma_{\text{LLM}}^2$ online, and the exponential gating $\exp(-\eta \cdot E_t)$ monotonically decreases weight as the estimated error grows.

\begin{remark}[Decay as a safety mechanism]
Time-based decay provides a \emph{worst-case guarantee}: regardless of LLM quality, the pseudo-observation weight vanishes as $t \to \infty$, ensuring that LinUCB converges to its standard regret bound asymptotically. Calibration-gated decay provides an \emph{adaptive} guarantee: the weight responds to the LLM's actual performance. The hybrid approach offers both.
\end{remark}

\section{Experimental Setup}
\label{sec:experiments}

\subsection{Environments}
\label{sec:environments}

We evaluate on two contextual bandit environments that test different aspects of LLM augmentation.

\paragraph{UCI Mushroom.} Following \citet{li2010contextual}, we construct a 2-arm bandit from the UCI Mushroom dataset~(8124 instances, 22 categorical features one-hot encoded to $d = 117$ dimensions). Arm~1 is ``eat'' and arm~2 is ``don't eat,'' with asymmetric rewards: eating an edible mushroom yields $+5$, eating a poisonous one yields $-35$, not eating an edible one yields $0$, and not eating a poisonous one yields $+5$. This creates a high-penalty exploration problem where incorrect predictions on poisonous mushrooms are very costly.

\paragraph{MIND-small.} We use the Microsoft News Dataset (MIND-small)~\citep{wu2020mind}, a news recommendation benchmark with 50{,}000+ impressions. Each impression provides a user with a click history and $K = 5$ candidate articles. We hash user history and article title/abstract into fixed-dimensional vectors using feature hashing, yielding per-arm context vectors. The reward is Bernoulli: $r = 1$ if the chosen article matches the clicked article, $0$ otherwise. Approximately 38\% of rounds have a clickable article among the candidates, making this a sparse-reward environment.

\begin{table}[t]
\centering
\caption{Environment comparison.}
\label{tab:environments}
\begin{tabular}{lcccc}
\toprule
& \textbf{Arms} & \textbf{Context dim} & \textbf{Reward range} & \textbf{LLM signal} \\
\midrule
Mushroom & 2 & 117 & $[-35, +5]$ & Low (raw codes) \\
MIND-small & 5 & per-arm & $\{0, 1\}$ & High (article titles) \\
\bottomrule
\end{tabular}
\end{table}

\subsection{Baselines and Configurations}
\label{sec:baselines}

All experiments use Disjoint LinUCB with $\alpha = 1.0$ and $\lambda_{\text{reg}} = 1.0$. We compare the following configurations:

\begin{table}[t]
\centering
\caption{Experimental configurations. All use GPT-4o-mini~(OpenAI) where an LLM is involved.}
\label{tab:configs}
\begin{tabular}{lllccc}
\toprule
\textbf{Name} & \textbf{Prompt} & \textbf{Decay} & $\lambda_w$ & $\eta$ & $\beta$ \\
\midrule
\textsc{NoLLM} & --- & zero & --- & --- & --- \\
\textsc{LLM-Default} & counterfactual & constant & 0.1 & --- & 0.95 \\
\textsc{LLM-Click} & mind\_click & constant & 0.1 & --- & 0.95 \\
\textsc{LLM-CalGated} & mind\_click & cal-gated & 0.3 & 10.0 & 0.95 \\
\textsc{LLM-Context} & counterfactual+ctx & constant & 0.1 & --- & 0.95 \\
\bottomrule
\end{tabular}
\end{table}

\textsc{LLM-Default} uses the generic counterfactual prompt (``predict what the unplayed arm would have given''). \textsc{LLM-Click} uses the task-specific \texttt{mind\_click} prompt (``predict click probability for each candidate article''). \textsc{LLM-CalGated} adds calibration-gated decay with $\eta = 10$, $\lambda_w = 0.3$ (higher ceiling since calibration gates it down). \textsc{LLM-Context} includes mushroom feature descriptions in the prompt history.

\subsection{Implementation Details}

All experiments run for $T = 100$ rounds with seed $= 42$ and $n_{\text{sims}} = 1$ (single simulation). The LLM is GPT-4o-mini via the OpenAI API, called through LangChain's \texttt{with\_structured\_output} for reliable JSON parsing. The \texttt{mind\_click} prompt style makes a single LLM call per round (predicting all arms simultaneously), while the default prompt requires two calls (calibration probe + scoring). Calibration tracking uses an EMA with $\beta = 0.95$.

We acknowledge that single-seed results without error bars are a limitation; these Phase~1 results establish directional findings that motivate targeted follow-up with multiple seeds and longer horizons.

\section{Results}
\label{sec:results}

\subsection{Mushroom: LLM Augmentation Hurts}
\label{sec:results-mushroom}

\begin{table}[t]
\centering
\caption{Cumulative regret on UCI Mushroom ($T = 100$, single seed).}
\label{tab:mushroom-results}
\begin{tabular}{lcccc}
\toprule
& $t=30$ & $t=50$ & $t=75$ & $t=100$ \\
\midrule
\textsc{NoLLM} & 125 & 170 & 190 & \textbf{265} \\
\textsc{LLM-Default} & 140 & 205 & 220 & 295 \\
\textsc{LLM-Context} & 180 & 205 & 305 & 365 \\
\bottomrule
\end{tabular}
\end{table}

\begin{figure}[t]
    \centering
    \includegraphics[width=0.7\textwidth]{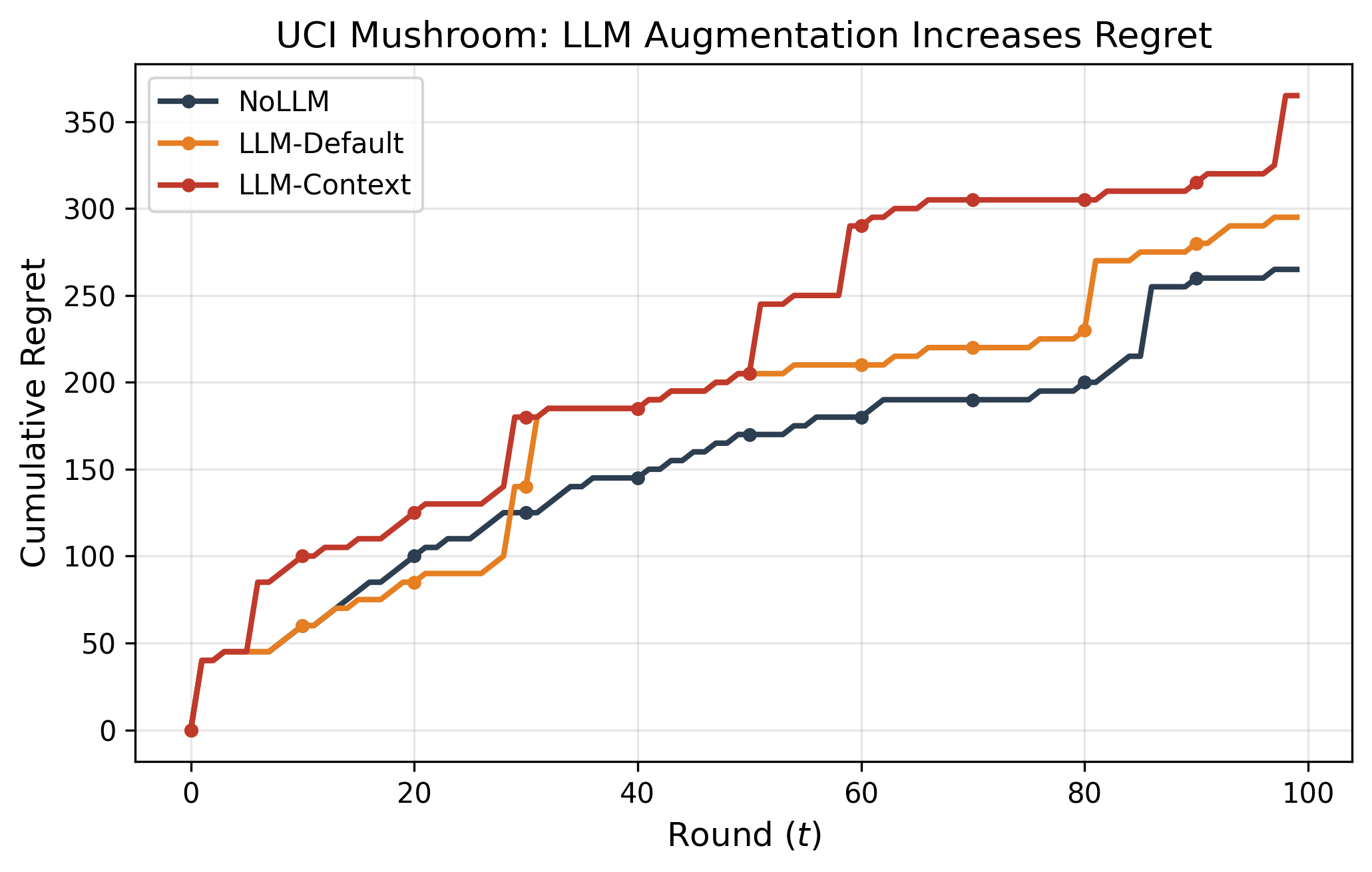}
    \caption{Cumulative regret on UCI Mushroom. LLM augmentation increases regret in all configurations.}
    \label{fig:mushroom-regret}
\end{figure}

On mushroom, \emph{all LLM variants increase regret} relative to the baseline (\Cref{tab:mushroom-results}, \Cref{fig:mushroom-regret}). The default prompt adds +30 cumulative regret at $t = 100$ (+11\%), while adding mushroom features to the prompt history (\textsc{LLM-Context}) adds +100 (+38\%).

The LLM's predictions on mushroom are inconsistent: it flips between predicting $-35$ and $+5$ for the same arm across rounds, because the context consists of raw categorical feature codes (e.g., \texttt{odor=foul}) that do not activate the LLM's pre-trained knowledge effectively. The constant decay at $\lambda_w = 0.1$ means this noise never stops influencing the learner.

\subsection{MIND: Task-Specific Prompting Reduces Regret}
\label{sec:results-mind}

\begin{table}[t]
\centering
\caption{Cumulative regret and token usage on MIND-small ($T = 100$, single seed).}
\label{tab:mind-results}
\begin{tabular}{lccccc}
\toprule
& $t=30$ & $t=50$ & $t=75$ & $t=100$ & Tokens (input) \\
\midrule
\textsc{NoLLM} & 10 & 17 & 22 & 32 & --- \\
\textsc{LLM-Default} & 11 & 18 & 26 & 35 & 163k \\
\textsc{LLM-Click} & \textbf{7} & \textbf{11} & \textbf{15} & \textbf{26} & 74k \\
\textsc{LLM-CalGated} & 9 & 15 & 22 & 33 & 74k \\
\bottomrule
\end{tabular}
\end{table}

\begin{figure}[t]
    \centering
    \includegraphics[width=0.7\textwidth]{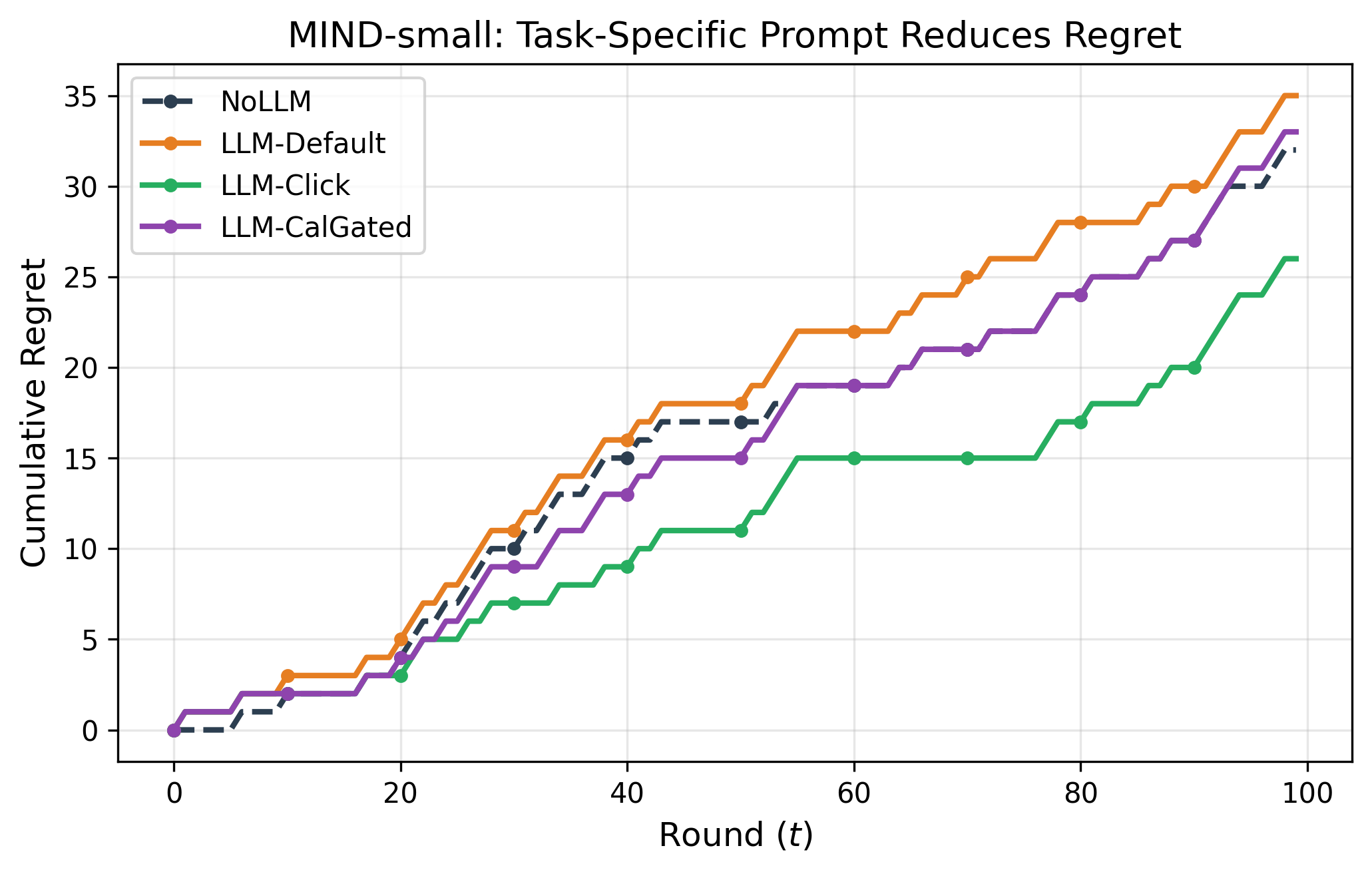}
    \caption{Cumulative regret on MIND-small. Only the task-specific prompt (\textsc{LLM-Click}) beats the baseline.}
    \label{fig:mind-regret}
\end{figure}

On MIND, \textsc{LLM-Click} is the only variant that consistently beats the baseline (\Cref{tab:mind-results}, \Cref{fig:mind-regret}). At $t = 100$, it achieves cumulative regret of 26 versus 32 for \textsc{NoLLM}---a 19\% reduction. The gap opens early ($t = 30$: 7 vs.\ 10) and widens throughout, suggesting that the LLM provides genuine value during the cold-start phase that compounds over time.

\textsc{LLM-Click} also uses 55\% fewer input tokens than \textsc{LLM-Default} (74k vs.\ 163k), because the task-specific prompt makes a single call per round predicting all arms simultaneously, rather than separate calibration and scoring calls.

\textsc{LLM-Default} slightly \emph{hurts} on MIND (35 vs.\ 32), demonstrating that the same underlying LLM model can help or hurt depending entirely on how the prediction task is framed.

\subsection{Ablation: Prompt Framing}
\label{sec:results-ablation-prompt}

The most striking finding across both environments is the dominance of prompt design. On MIND, holding the algorithm and decay schedule constant, switching from the generic counterfactual prompt to the task-specific \texttt{mind\_click} prompt changes the outcome from +3 regret (worse than baseline) to $-6$ regret (better than baseline)---a swing of 9 points entirely attributable to prompt framing.

The mechanism is straightforward: the generic prompt asks the LLM to reason about ``what reward would arm~$a$ have produced?''---an abstract counterfactual question that it answers with hedged, mid-range predictions (0.2--0.5 for binary click probabilities). The task-specific prompt asks ``will this user click this article?''---a concrete prediction task that aligns with the LLM's pre-training on natural language understanding and enables it to leverage its knowledge of user preferences and article content.

\subsection{Ablation: Calibration-Gated Decay}
\label{sec:results-ablation-cal}

Calibration-gated decay with $\eta = 10$ (\textsc{LLM-CalGated}) performs \emph{worse} than the simple constant schedule (\textsc{LLM-Click}), achieving regret of 33 versus 26 (\Cref{tab:mind-results}).

The failure mode is instructive. On MIND, the LLM's calibration errors are consistently small---the EMA $E_t$ stabilizes around 0.05--0.08. With $\eta = 10$, this yields $\exp(-10 \times 0.06) \approx 0.55$, causing the effective weight to decay from $0.3 \times 0.55 \approx 0.17$ to below 0.02 by $t = 100$. The gating is too aggressive: it silences the LLM even though its predictions are directionally useful.

This reveals a fundamental tension in calibration gating: when the LLM is \emph{reasonably good} (small errors), aggressive gating removes useful signal. The $\eta$ parameter must be tuned per domain, and the regime where calibration gating adds value over a simple constant schedule may be narrow---it requires that the LLM's accuracy varies meaningfully across rounds, so that gating can selectively amplify good rounds.

\section{Discussion}
\label{sec:discussion}

\subsection{When Do LLMs Help?}

Our results delineate a clear condition for LLM augmentation to reduce regret: the LLM must have \emph{genuine domain knowledge} relevant to the prediction task, and the prompt must be designed to \emph{activate} that knowledge.

On MIND, the LLM can read article titles (``Tech Giants Report Record Earnings''), user reading history (previously clicked technology and business articles), and make informed click predictions. This is a task that aligns naturally with the LLM's pre-training on text understanding. On mushroom, the LLM receives raw categorical codes (\texttt{cap-shape=convex}, \texttt{odor=foul}) that do not activate useful pre-trained knowledge---the LLM ``knows'' that foul odor suggests danger, but cannot reliably map this to the specific reward structure ($-35$ vs.\ $+5$) without in-context learning that the prompt history cannot efficiently support.

\subsection{Prompt Framing as the Bottleneck}

The dominance of prompt framing over algorithmic tuning is our most practically relevant finding. Researchers and practitioners designing LLM-augmented bandit systems should invest primarily in prompt engineering---crafting prompts that frame the prediction task in terms the LLM can reason about naturally---rather than in sophisticated decay schedule design. The decay schedule provides a safety mechanism (eventually silencing the LLM), but the prompt determines whether the LLM provides useful signal in the first place.

\subsection{Limitations}

Our results are subject to several important limitations:

\begin{enumerate}
    \item \textbf{Single seed.} All results use a single random seed ($= 42$) with one simulation. Error bars and statistical significance tests require $n_{\text{sims}} \geq 5$.
    \item \textbf{Short horizon.} At $T = 100$, we cannot determine whether the LLM's advantage persists, widens, or reverses as LinUCB accumulates more data. Runs of 300--500 rounds would clarify the crossover dynamics.
    \item \textbf{Single LLM.} We test only GPT-4o-mini. Different LLMs may exhibit different calibration profiles and domain knowledge.
    \item \textbf{Two environments.} Broader evaluation across additional domains (e.g., advertising, clinical trials) is needed.
    \item \textbf{Cost.} Even with the efficient single-call \texttt{mind\_click} prompt (74k tokens over 100 rounds), LLM calls add latency and monetary cost per round. The cost--regret trade-off has not been formalized.
    \item \textbf{Confidence unused.} The LLM returns per-prediction confidence scores that we do not currently use for weighting.
\end{enumerate}

\subsection{Future Work}

Several extensions follow naturally:
\begin{itemize}
    \item \textbf{Longer horizons and multiple seeds} to establish statistical significance and crossover dynamics.
    \item \textbf{Gentler calibration gating} ($\eta \in [1, 3]$) to preserve LLM signal while penalizing genuinely poor rounds.
    \item \textbf{Confidence-weighted injection}, scaling pseudo-observation weight by the LLM's self-reported confidence.
    \item \textbf{Sweep over $\lambda_w$} to find the optimal base weight for the cold-start phase.
    \item \textbf{Task-specific prompts for mushroom} (``is this mushroom safe to eat?'') to test whether the prompt framing insight transfers.
    \item \textbf{Formal regret analysis} deriving bounds on the additional regret (or regret reduction) from pseudo-observations as a function of LLM bias and weight schedule.
\end{itemize}

\section{Conclusion}
\label{sec:conclusion}

We have presented a framework for augmenting contextual bandits with LLM pseudo-observations, where the injection weight is controlled by calibration-gated decay schedules. On MIND-small news recommendation, task-specific LLM prompting reduces cumulative regret by 19\% relative to pure LinUCB. On UCI Mushroom, where the LLM lacks actionable domain knowledge, augmentation increases regret regardless of prompt design or decay schedule.

Our central finding is that \emph{prompt framing is the dominant factor} in LLM-augmented bandit performance---more important than the choice of decay schedule, calibration gating parameters, or even whether calibration gating is used at all. This suggests that the most productive direction for practitioners is to invest in task-specific prompt design that activates the LLM's pre-trained domain knowledge, while using simple constant decay as a sufficient algorithmic mechanism.

\bibliographystyle{plainnat}
\bibliography{references}

@inproceedings{li2010contextual,
  title     = {A Contextual-Bandit Approach to Personalized News Article Recommendation},
  author    = {Li, Lihong and Chu, Wei and Langford, John and Schapire, Robert E.},
  booktitle = {Proceedings of the 19th International Conference on World Wide Web (WWW)},
  pages     = {661--670},
  year      = {2010},
}

@article{abbasi2011improved,
  title   = {Improved Algorithms for Linear Stochastic Bandits},
  author  = {Abbasi-Yadkori, Yasin and P{\'a}l, D{\'a}vid and Szepesv{\'a}ri, Csaba},
  journal = {Advances in Neural Information Processing Systems},
  volume  = {24},
  year    = {2011},
}

@book{lattimore2020bandit,
  title     = {Bandit Algorithms},
  author    = {Lattimore, Tor and Szepesv{\'a}ri, Csaba},
  year      = {2020},
  publisher = {Cambridge University Press},
}

@inproceedings{liu2023bypassing,
  title     = {Bypassing the Simulator: Near-Optimal Adversarial Linear Contextual Bandits},
  author    = {Liu, Haolin and Wei, Chen-Yu and Zimmert, Julian},
  booktitle = {Advances in Neural Information Processing Systems (NeurIPS)},
  year      = {2023},
}

@inproceedings{alamdari2024jump,
  title     = {Jump Starting Bandits with {LLM}-Generated Prior Knowledge},
  author    = {Alamdari, Parand A. and Cao, Yanshuai and Wilson, Kevin H.},
  booktitle = {Proceedings of the 2024 Conference on Empirical Methods in Natural Language Processing (EMNLP)},
  pages     = {19821--19833},
  year      = {2024},
  publisher = {Association for Computational Linguistics},
}

@article{baheri2023llm,
  title   = {{LLMs}-augmented Contextual Bandit},
  author  = {Baheri, Ali and Alm, Cecilia O.},
  journal = {arXiv preprint arXiv:2311.02268},
  year    = {2023},
}

@article{sun2025llm,
  title   = {Large Language Model-Enhanced Multi-Armed Bandits},
  author  = {Sun, Jiahang and Wang, Zhiyong and Yang, Runhan and Xiao, Chenjun and Lui, John C. S. and Dai, Zhongxiang},
  journal = {arXiv preprint arXiv:2502.01118},
  year    = {2025},
}

@article{ye2024lola,
  title   = {{LOLA}: {LLM}-Assisted Online Learning Algorithm for Content Experiments},
  author  = {Ye, Zikun and Yoganarasimhan, Hema and Zheng, Yufeng},
  journal = {arXiv preprint arXiv:2406.02611},
  year    = {2024},
}

@inproceedings{hazime2025evaluation,
  title     = {Evaluation of {LLM} Powered Agentic {AI} for Solving Multi-Arm Bandit Problems},
  author    = {Hazime, Jawad and Farooq, Junaid},
  booktitle = {2025 IEEE International Conference on Omni-layer Intelligent Systems (COINS)},
  year      = {2025},
  publisher = {IEEE},
}

@inproceedings{chen2024efficient,
  title     = {Efficient Sequential Decision Making with Large Language Models},
  author    = {Chen, Dingyang and Zhang, Qi and Zhu, Yinglun},
  booktitle = {Proceedings of the 2024 Conference on Empirical Methods in Natural Language Processing (EMNLP)},
  pages     = {9157--9170},
  year      = {2024},
  publisher = {Association for Computational Linguistics},
}

@inproceedings{nie2025evolve,
  title     = {{EVOLvE}: Evaluating and Optimizing {LLMs} For In-Context Exploration},
  author    = {Nie, Allen and Su, Yi and Chang, Bo and Lee, Jonathan N. and Chi, Ed H. and Le, Quoc V. and Chen, Minmin},
  booktitle = {Proceedings of the 42nd International Conference on Machine Learning (ICML)},
  series    = {PMLR},
  volume    = {267},
  year      = {2025},
}

@inproceedings{huch2024rome,
  title     = {{RoME}: A Robust Mixed-Effects Bandit Algorithm for Optimizing Mobile Health Interventions},
  author    = {Huch, Easton and Shi, Jieru and Abbott, Madeline R. and Golbus, Jessica R. and Moreno, Alexander and Dempsey, Walter H.},
  booktitle = {Advances in Neural Information Processing Systems (NeurIPS)},
  year      = {2024},
}

@inproceedings{wu2020mind,
  title     = {{MIND}: A Large-scale Dataset for News Recommendation},
  author    = {Wu, Fangzhao and Qiao, Ying and Chen, Jiun-Hung and Wu, Chuhan and Qi, Tao and Lian, Jianxun and Liu, Dongsheng and Xie, Xing and Gao, Jianfeng and Wu, Winnie and Zhou, Ming},
  booktitle = {Proceedings of the 58th Annual Meeting of the Association for Computational Linguistics (ACL)},
  pages     = {3597--3606},
  year      = {2020},
}

@inproceedings{guo2017calibration,
  title     = {On Calibration of Modern Neural Networks},
  author    = {Guo, Chuan and Pleiss, Geoff and Sun, Yu and Weinberger, Kilian Q.},
  booktitle = {Proceedings of the 34th International Conference on Machine Learning (ICML)},
  pages     = {1321--1330},
  year      = {2017},
}

@article{kadavath2022language,
  title   = {Language Models (Mostly) Know What They Know},
  author  = {Kadavath, Saurav and Conerly, Tom and Askell, Amanda and Henighan, Tom and Drain, Dawn and Perez, Ethan and Schiefer, Nicholas and Hatfield-Dodds, Zac and DasSarma, Nova and Tran-Johnson, Eli and others},
  journal = {arXiv preprint arXiv:2207.05221},
  year    = {2022},
}

\appendix

\section{Experimental Details}
\label{app:experiment-details}

\subsection{Hyperparameters}

All experiments use the following shared hyperparameters: exploration parameter $\alpha = 1.0$, ridge regularization $\lambda_{\text{reg}} = 1.0$, calibration EMA parameter $\beta = 0.95$, random seed $= 42$, single simulation. MIND experiments use $K = 5$ arms with hashed context vectors. Mushroom experiments use $K = 2$ arms with one-hot encoded categorical features ($d = 117$ after bias augmentation).

\subsection{Prompt Templates}

\paragraph{Generic counterfactual prompt (default).} The prompt provides the current context description, lists available arms with descriptions, includes a history of the last 5 rounds (arm chosen, reward observed), and asks: ``For each unplayed arm, predict the expected reward the learner would have received. Return predictions as JSON.'' The calibration probe variant asks for the played arm's expected reward \emph{before} the outcome is revealed.

\paragraph{Task-specific prompt (mind\_click).} The prompt provides the user's recent reading history (last 10 article titles), lists 5 candidate articles with titles, and asks: ``For each candidate article, predict the probability that this user will click on it. Return predictions as JSON with predicted\_reward in [0, 1].'' A single call covers all arms, serving as both calibration probe and scoring.

\subsection{Compute Budget}

All LLM experiments use GPT-4o-mini via the OpenAI API. Token usage per 100-round experiment: \textsc{LLM-Default} consumes ${\sim}163$k input tokens; \textsc{LLM-Click} consumes ${\sim}74$k input tokens. Wall-clock time is dominated by API latency (${\sim}0.5$--$1$s per call).

\section{Additional Results}
\label{app:additional-results}

\subsection{Calibration Error Trajectories}

The calibration EMA $E_t$ on MIND stabilizes between 0.05 and 0.08 for both \textsc{LLM-Click} and \textsc{LLM-CalGated}, indicating that the LLM's click predictions are consistently close to actual outcomes. This small and stable error is precisely why aggressive calibration gating ($\eta = 10$) fails---the gating function cannot distinguish between ``good enough'' and ``perfect,'' and treats both as grounds for weight reduction.

On mushroom, calibration errors are larger and more variable, reflecting the LLM's inconsistent predictions on raw feature codes. In this regime, calibration gating would correctly reduce weight---but since the underlying predictions are already harmful, gating alone is insufficient; the prompt framing must be fixed first.

\subsection{MIND Tiny Results}

Sanity-check experiments with $T = 10$ rounds show regret of 1--2 for all configurations, consistent with the sparse-reward structure (only ${\sim}38$\% of rounds have a clickable article). These runs verified the pipeline but are too short to differentiate configurations.

\end{document}